\documentclass[11pt,a4paper]{scrartcl}
\usepackage[sort&compress,square,comma,authoryear]{natbib}

\usepackage[top=3cm, bottom=3cm, left=3cm, right=3cm]{geometry}

\usepackage{booktabs}

\usepackage{color}
\usepackage{latexsym}              
\usepackage{amsmath}               
\usepackage{amssymb}               
\usepackage{amsfonts}              
\usepackage{amsthm}                
\usepackage{multirow}
\usepackage{tikz}                  
\usetikzlibrary{arrows,positioning,shapes}
\usepackage{tcolorbox}
\usepackage{algorithm2e}

\usepackage[english]{babel}
\usepackage{fontawesome}

\RequirePackage[%
  pdfstartview=FitH,%
  breaklinks=true,%
  bookmarks=true,%
  colorlinks=true,%
  linkcolor= blue,
  anchorcolor=blue,%
  citecolor=blue,
  filecolor=blue,%
  menucolor=blue,%
  urlcolor=blue%
  ]{hyperref}
  
  \AtBeginDocument{%
  \hypersetup{%
    pdfauthor={},%
    	colorlinks = true,%
  	urlcolor = blue,%
  	linkcolor = blue,%
  	citecolor = orange,%
    pdftitle={}%
  }
}

\newcommand{\N}{\mathbb{N}}

\usepackage[mathcal]{eucal}
\usepackage{cleveref}
\crefname{assumption}{Assumption}{Assumptions}
\crefname{equation}{Eq.}{Eqs.}
\crefname{figure}{Fig.}{Figs.}
\crefname{table}{Table}{Tables}
\crefname{section}{Sec.}{Secs.}
\crefname{theorem}{Thm.}{Thms.}
\crefname{lemma}{Lemma}{Lemmas}
\crefname{corollary}{Cor.}{Cors.}
\crefname{example}{Example}{Examples}
\crefname{appendix}{Appendix}{Appendixes}
\crefname{remark}{Remark}{Remark}

\makeatother

\theoremstyle{plain}  
\newtheorem{theorem}{Theorem}

\newtheorem{lemma}[theorem]{Lemma}
\newtheorem{proposition}[theorem]{Proposition}

\begin{document}
\title{Revisiting clustering as matrix factorisation on the Stiefel manifold}

\author{\textbf{St\'ephane Chr\'etien} \\ [2ex]
Universit\'e Lumi\` ere-Lyon-II \\
The Alan Turing Institute, London, UK \\
The National Physical Laboratory, UK \\
\url{stephane.chretien@univ-lyon2.fr} \\\\
\textbf{Benjamin Guedj} \\ [2ex]
Inria, Lille - Nord Europe research centre, France \\
University College London, Department of Computer Science\\ and Centre for Artificial Intelligence, UK \\
\url{benjamin.guedj@inria.fr}
}
\date{}

\maketitle

\begin{abstract}
This paper studies clustering for possibly high dimensional data (\emph{e.g.} images, time series, gene expression data, and many other settings), and rephrase it as low rank matrix estimation in the PAC-Bayesian framework. Our approach leverages the well known Burer-Monteiro factorisation strategy from large scale optimisation, in the context of low rank estimation. Moreover, our Burer-Monteiro factors are shown to lie on a Stiefel manifold. We propose a new generalized Bayesian estimator for this problem and prove novel prediction bounds for clustering. We also devise a componentwise Langevin sampler on the Stiefel manifold to compute this estimator. 
\end{abstract}


\section{Introduction}
   Clustering, \emph{i.e.,} unsupervised classification, is a central problem in machine learning and has attracted great attention since the origins of statistics, via model-based learning, but recently regained a lot of interest from theoreticians, due to its similarities with community detection \cite{arias2014community,verzelen2015community}. On the application side, clustering is pervasive in data science, and has become a basic tool in computer science, bio-informatics, finance, metrology, to name but a few.
  
  \subsection{Historical background}
  The problem of identifying clusters in a data set can be addressed using an wide variety of tools. Two main approaches can be delineated, namely the model-based approach and the non-model based approach. Techniques such as hierarchical clustering \cite{hastie2009unsupervised}, minimum spanning tree-based approaches \cite{blum2016foundations}, $K$-means algorithms \cite{hastie2009unsupervised}, belong to the non-model based family of methods. Model-based techniques mostly rely on mixture modelling \cite{mclachlan2004finite} and often offer better interpretability whilst being easily amenable to uncertainty quantification analysis. The EM algorithm \cite{dempster1977maximum,mclachlan2004finite} is often the algorithm of choice in the frequentist approach while many Monte Carlo Markov Chain techniques have been proposed for estimation in the Bayesian setting.
  
  In recent years, the clustering problem has revived a surge of interest in a different setting, namely community detection in random graphs. Tools from spectral graph theory and convex optimisation, combined with recent breakthrough from random matrix theory were put to work in devising efficient clustering methods that operate in polynomial time. The celebrated example of  Max-Cut, a well known NP-hard combinatorial optimisation problem strongly related to bi-clustering and with a tight Semi-Definite Programming (SDP) relaxation discovered by \cite{goemans1995improved}, is an example among the many successes of the convex optimisation approach to addressing machine learning problems.
  SDP is the class of optimisation problems that consist in minimising a linear function over the sets of Positive Semi-Definite matrices that satisfy a set of linear (in)equalities. \cite{goemans1995improved} subsequently triggered a long lasting trend of research in convex relaxation with many application in data science, and recent results proposing tighter relaxations to the clustering problem can be found in \cite{guedon2016community}, \cite{chretien2016semi}, \cite{giraud2018partial}. Some of these methods even apply to any kind of data set endowed with a relevant affinity measure computed from the pairwise distances between the data points, and share the common feature of using low-rank matrix representations of the clustering problem. The theoretical tools behind the analysing of the performance of these convex optimisation-based methods are also quite fascinating and range from random matrix theory \cite{bandeira2018random,vershynin2018high},  concentration inequalities for quadratic forms of random vectors \cite{rudelson2013hanson} and optimisation theory  \cite[optimality conditions, see][]{royer2017adaptive}, localisation arguments in statistical learning theory \cite{giraud2018partial}, Grothendieck's inequality \cite{guedon2016community,montanari2015semidefinite}, to name but a few.
  
  The main drawback, however, of the current lines of approach to the performance analysis of these powerful convex \textit{SDP} and \textit{spectral relaxations} is that they all depend on the separation between clusters, \emph{i.e.,} the minimum distance between two points from different clusters, a crucial parameter in the aforecited analyses. In real data sets however, sufficient inter-cluster separation rarely holds and overlaps between clusters are the common situation. This leaves open the difficult problem of finding an alternative theoretical route for controlling the estimation error. On the computational side, the sample size is also a problem for SDP relaxations for which off-the-shelf software does not scale to big data. A remedy to this problem is to use the Burer-Monteiro  factorisation consisting in solving in $U$ where $X=UU^t$ is the variable of the SDP at hand \cite{burer2003nonlinear}. The Burer-Monteiro factorisation results in a non-convex optimisation problem whose local minimisers are global minimisers when the number of columns of $U$ is sufficiently large \cite{burer2005local,boumal2016non}. In practice however, the rank of the sought matrix is simply equal to the number of clusters, and whether such small priors on the rank of the Burer-Monteiro factorisation are compatible with the local/global equivalence of the minimisers in general remains an open question to this day. A final source of frustration in our list, is that there does not seem to exist any method for quantifying the uncertainty of the results in these convex optimisation-based approaches to clustering. 
  
  In the present paper, we propose a generalized Bayesian approach to clustering which hinges on low rank estimation of a clustering matrix. We then leverage arguments from the PAC-Bayesian theory for controlling the error which does not use any prior estimate of separation. Our approach is based on the estimation of a normalised version $T^*$ of the adjacency matrix of the clustering, which can be factorised into $T^*=U^*U^{*^t}$, where $U^*$ has orthonormal, non-negative columns. Leveraging this structure leads to sampling on the intersection of the Stiefel manifold \cite{edelman1998geometry} and the non-negative orthant, which is another surprising manifestation of the power of non-negative matrix factorisation (NMF) in clustering problems. Solving this factorised version in the PAC-Bayesian setting is the sampling counterpart of the Burer-Monteiro approach to the numerical solution of high dimensional SDP. The PAC-Bayesian approach (initiated by \cite{ShaweTaylorW97,McAllester98,McAllester99,catoni2004statistical,Catoni07}; see \cite{guedj2019primer} for a recent survey) moreover makes no prior use of the separation and at the same time makes it possible to obtain state-of-the-art risk bounds.

  \subsection{Our contribution}
  The main goal of the present paper is to study the clustering problem from a low rank Stiefel matrix, i.e. matrices with orthonormal columns, view point, and present a PAC-Bayesian analysis of the related statistical estimation problem. Our approach is in particular inspired by recent work on low rank approximation for $k$-means \cite{boutsidis2009unsupervised,cohen2015dimensionality}, where the representation of clustering using the matrix $T^*$ is explicitly stated (although no algorithm is provided), and PAC-Bayesian bounds for Non-Negative Matrix factorisation \cite[as introduced by][although they do not establish the link between NMF and clustering]{alquier2017oracle}. To the best of our knowledge, the representation in \cite{boutsidis2009unsupervised} using the matrix $T^*$ has never been studied from a statistical  learning perspective.
  
  We present our main result (Theorem \ref{main}, which states an inequality holding in expectation on the prediction performance) in Section \ref{sec:main}. Our second main result is Theorem \ref{incoh}, which specifies the results of Theorem \ref{main} in the case where we assume that the family of means is incoherent. Section \ref{sec:algo} is devoted to our algorithm (an alternating Langevin sampler which relies on computing gradients on the Stiefel manifold). 

Our choice of the Langevin sampler is motivated by its faster convergence than Metropolis type samplers due to the absence of any rejection step. This is even more true in the case where one needs to account for the manifold structure of the model as is the case in the present paper.

  Additional proofs are gathered in the supplementary material.
  
  \subsection{Notation}
  The notation used in the present paper is fairly standard. The canonical scalar product in $\mathbb R^d$ will be denoted by $\langle\cdot,\cdot\rangle$, the $\ell_p$ norms by $\Vert \cdots\Vert_p$. For matrices in $\mathbb R^{d\times n}$, the operator norm will be denoted by $\Vert \cdot \Vert$ and the Frobenius norm by $\Vert \cdot \Vert_F$. The Stiefel manifold of order $(n,R)$, i.e. the set of matrices in $\mathbb R^{n\times R}$ with orthonormal columns, will be denoted by $\mathbb O_{n,R}$, and $\mathbb O_{n,R,+}$ will denote the subset of the Stiefel manifold $\mathbb O_{n,R}$ consisting of componentwise nonnegative matrices. The matrices in $\mathbb O_{n,R}$ will sometimes be identified with matrices in $\mathbb R^{n\times n}$ where the first $R$ columns form an orthonormal family and the remaining $n-R$ columns are set to zero. The gradient operator acting on differentiable multivariate functions will be denoted by $\nabla$.
  
  \section{Non-negative factorisation of the Stiefel manifold}\label{sec:main}
  
  This section is devoted to the presentation of our framework and our main theoretical result.
  
  \subsection{Model}
  Let data points $x_1,\ldots,x_n$ be vectors in $\mathbb R^d$ and let $X$ denote the matrix
  \begin{equation*}
  X  =  
  \left[
  x_1,\ldots,x_n
  \right].
  \end{equation*}
  Let $\mu_1,\ldots,\mu_K$ be $K\in\N\backslash\{0\}$ vectors in $\mathbb R^d$. 
  We will say that $x_i$ belongs to cluster $k\in\{1,\dots,K\}$ if $x_i=\mu_k+E_i$ for some centered random vector $E_i\in \mathbb R^d$. 
  For each $i=1,\ldots,n$, we will denote by $k_i$ the label of the cluster to which $x_i$ belongs. For each $k$, 
  we will denote by $I_k$ the index set of the points which belong to cluster $k$ and $n_k$ its cardinality. Now, we can decompose $X$ as $X=M+E$ with  
  \begin{equation*}
  M  =  \left[\mu_{k_1},\ldots,\mu_{k_n}\right],
  \end{equation*}
  \begin{equation*}
  E  =  \left[\epsilon_{1},\ldots,\epsilon_{n}\right]
  \end{equation*}
    More assumptions about the noise matrix $E$ will be introduced and their consequences on the performance of our clustering method will be studied in Theorem \ref{incoh}.
    
  \subsection{Ideal solution}
  If we try to estimate the columns of $M$, one simple way is to use a convex combination of the $x_i$'s for 
  each of them. In other words, one might try to approximate $X$ by $XT^*$ where $T^*$ is a $\mathbb R^{n\times n}$ 
  matrix. One simple way to proceed is to set $T$ as the matrix which computes the cluster means, given by
  \begin{align*}
  T^*_{i,j} & =
  \begin{cases}
  \: \frac1{n_{k}} &\textrm{ if $x_i$ and $x_j$ belong to the same cluster $k$, \emph{i.e.}, $k_i=k_j$} \\
  \\
  \: 0 &\textrm{ otherwise.} 
  \end{cases}
  \end{align*}
  Thus, each column $i=1,\ldots,n$ of $XT^*$ is simply the mean over cluster $k_i$. This type of solution is well-motivated 
  by the fact that the mean is the least-squares solution of the approximation of $\mu_k$ by the observation points. The matrix $T^*$ defined as above enjoys the following desirable properties: (i) its rank is exactly the number of clusters (ii) it is nonnegative (iii) the columns corresponding to different clusters are orthogonal.

  One important fact to notice is that the eigenvalue decomposition of $T^*$ is explicit and given by 
  \begin{equation}
  T^*  =  U^*  {U^*}^t
  \label{Talph}
  \end{equation}
  with 
  \begin{equation}
  U^*  = \left[ \frac1{\sqrt{n_1}} \ 1_{I_{1}},\dots, \frac1{\sqrt{n_K}} 1_{I_{K}}\right], 
  \label{UVbar}
  \end{equation}
  and therefore, all the eigenvalues of $T^*$ are equal to one. 
  
  Based on this decomposition, we can now focus on estimating $U^*$ rather than $T^*$, the reason being that working on estimating $U^*$ with $\hat U\ge 0$ will automatically enforce positive semi-definiteness of $\hat T$ (the estimator of $T^*$) and non-negativity of its components. Moreover, enforcing the orthogonality of the columns of $\hat U$, combined with  the non-negativity of its components, will enforce the columns of $\hat U$ to have disjoint supports.
  
  Adopting a generalized Bayesian strategy \cite[inspired by][]{alquier2017oracle}, we will then define a prior distribution on $\hat U$ and study the main properties of the resulting (generalized) posterior distribution.
  
  \subsection{The latent variable model}
  In order to perform an accurate estimation, we need to devise 
  meaningful priors which will account for the main constraints our estimator should satisfy, namely (i) nonnegativity of the entries
  (ii) orthogonality of the columns
  (iii) the columns have unit $\ell_2$ norm
  (iv) group sparsity of the columns.

  In order to simplify this task, we will introduce a \textit{latent (matrix) variable $O$} with uniform distribution on the orthogonal group, and \textit{build priors on $U$} that will promote group sparsity of the columns and non-negativity (component-wise). 
  
  \paragraph{Prior on $(U,O)$.}
  Let $\mathcal U_R$ denote the set of matrices of the form 
  \begin{align*}
  U & = 
  \begin{bmatrix}
  U_{1,1} & \cdots & U_{1,R} & 0 & \cdots & 0 \\
  \vdots  &        & \vdots  & \vdots & & \vdots \\
  U_{n,1} & \cdots & U_{n,R} & 0 & \cdots & 0 
  \end{bmatrix}.
  \end{align*}
  Let $\mathbb O_{n,R}$ denote the Stiefel manifold, \emph{i.e.,} the manifold of all matrices with $R$ orthonormal columns in $\mathbb R^n$.
  The prior on $(U,O)\in \mathcal U_R\times \mathbb O_{n,R}$ is given by
  \begin{align*}
  \pi_{U,O}(U,O) & = \pi_{U\mid O}(U) \ \pi_O(O)
  \end{align*} 
  with 
  \begin{align*}
  \pi_{U_{i,r}\mid O} (U) & = \frac{1}{\sqrt{2\pi}\mu}\exp \left(-\frac{ ( U_{i,r}- \vert O_{i,r}\vert )^2}{2\mu^2}\right),
  \end{align*}
for $i=1,\ldots,n$, and $r=1,\ldots,R$, with $R$ being a fixed integer and $\pi_O$ being the uniform distribution on the Stiefel manifold $\mathbb O_{n,R}$.  
  \subsection{Generalized posterior and estimator}
  
  Following the approach of \cite{alquier2017oracle}, we use a loss term \cite[instead of a likelihood, hence the term "generalized Bayes", see][for a survey]{guedj2019primer} given by
  \begin{align*}
  L_\lambda(U) & = \exp \left(-\frac{\lambda}2 \ \Vert X-XUU^t\Vert_F^2\right)
  \end{align*}
  for some fixed positive parameters $\lambda$ and $\mu$. The resulting generalized posterior (also known as a \emph{Gibbs measure}) is defined as
  \begin{align*}
  \rho(U,O) & = \frac1{Z_\lambda} \ L_\lambda(U) \ \pi_{U\mid O} (U) \ 
  \pi_O(O),
  \end{align*}
  where $Z_\lambda$ denotes the normalisation constant
  $
  Z_\lambda  = \int \ L_\lambda(U) \ \pi_{U\mid O}(U) \ \pi_{O}(O)  \  dU.
  $
  Finally we let $\hat U_{\lambda}$ denote the posterior mean of $U$, i.e. 
  \begin{align*}
      \hat U_\lambda & = 
      \int U \ L_\lambda(U) \ \pi_{U\mid O}(U) \pi_O(O) dU dO.
  \end{align*}

  \subsection{A PAC-Bayesian-flavored error bound}
  
  Our main result is the following theorem. 
  \begin{theorem}    
    \label{main}
        Let us assume that $E$ is fixed. Let $\nu_{\min}$  and $\nu_{\max}$ be such that 
    \begin{align*}
    \nu_{\min} \le \min_{\tilde U \in \mathbb O_{n,R,+}, \ M\tilde U \tilde U^t=M}\quad \Vert E(I-\tilde U \tilde U^t)\Vert_F
    \end{align*}
    and
    \begin{align*}
    \nu_{\max} \ge \max_{\tilde U \in \mathbb O_{n,R,+}, \ M\tilde U \tilde U^t=M}\quad \Vert E(I-\tilde U \tilde U^t)\Vert_F.
    \end{align*}
    Then, for all $\epsilon>0$, and for all $c_O> 0$  and $c_U>c_O$ such that 
    \begin{equation}
    c_U  \left(2+c_U\right)  \le \epsilon \frac{\nu_{\min}}{\Vert M\Vert+\Vert E \Vert},
    \label{epscu}
    \end{equation}
    for any $R \in \{1,\ldots,n\}$ and for $c$ and $\rho$ sufficiently small universal constants, 
    we have 
    \begin{multline}
    \mathbb E \left[\left\Vert M \left(T^*-\hat U_{\lambda}\hat U_{\lambda}^t\right)\right\Vert_F\right] 
    \le  (1+\epsilon) \min_{\tilde U \in \mathbb O_{n,R,+}, \ M\tilde U \tilde U^t=M}\quad \Bigg\{ \Vert M(T^*-\tilde U\tilde U^{t})\Vert_F \\ 
    +\sqrt{\frac{1}
      {\exp \left(\frac{\left(\sqrt{c_U^2-c_O^2}-\sqrt{nR} \right)^2}{2} \right) - 1}} \\
      +\sqrt{\left(nR-\frac12 (R^2+R)\right) \log(\rho^{-1})+\log(c^{-1})} +(2+\epsilon)\nu_{\max}\Bigg\}.
      \label{riskbnd}
    \end{multline} 
  \end{theorem}  
  This theorem gives a prediction bound on the difference between the true and the estimated cluster matrices filtered by the matrix of means. 
  Up to our knowledge, this is the first oracle bound for clustering using a generalized Bayesian NMF.
  Note that the oracle bound is not sharp as the leading constant is $1+\epsilon>1$, however $\epsilon$ may be chosen arbitrarily close to $0$.
  
  Note also that the claim that this result is PAC-Bayesian-flavored comes from the fact that the prediction machinery is largely inspired by \cite{alquier2017oracle}, and the scheme of proof builds upon the PAC-Bayesian bound from \cite{dalalyan2008aggregation}. Hence we kept that PAC-Bayesian filiation, even though the bound holds in expectation.
  
  The following Theorem gives a more precise bound in the case where the noise $E$ is assumed to be iid Gaussian.
  \begin{theorem}\label{incoh}
    Assume that the dimension is larger that the number of clusters, i.e. $d>K$. Fix $\nu_{\max}>\nu_{\min}>0$ \footnote{as the reader will be able to check, the values of $\nu_{\max}$ and $\nu_{\min}$ will play an essential role in the expression of the success probability of the method.}. 
    In addition to the assumptions of Theorem \ref{main}, assume that 
    $E$ is iid Gaussian with minimum (resp. maximum) one-dimensional variance $\sigma_{\min}^2$ (resp. $\sigma_{\max}^2$) and assume also that the $\mu_k$ have Euclidean norm less that $1$ and pairwise scalar products less than $\mu$ in absolute value. Then, as long as $\mu <1/(K-1)$, for all $\epsilon>0$, and for all $c_O> 0$  and $c_U>c_O$ such that 
    \begin{multline}
        c_U  \left(2+c_U\right) \nonumber \le 
        \epsilon \frac{\nu_{\min}}{\sqrt{\left(\max_{k=1}^K \ n_k\right) \         \mu(K-1)+1}+\sigma_{\max} \left(\sqrt{n}+2\sqrt{d}\right)},
    \label{epscuincoh}
    \end{multline}
    with probability at least 
    $$1-\exp(-d)-\exp (-nu^2/8)- K,
    $$  
    with 
    \begin{align}
        K & =\left(\frac{c}{\epsilon}\right)^{nR-R(R+1)/2} \ \left(\frac{2}{\sqrt{\pi n(n-R)}} \left(t_{\min} \ e/2 \right)^{n(d-R)/4} +\exp(-t_{\max})\right),
    \end{align}
    we have
    \begin{multline*}
            \sum_{k=1}^K \sum_{i_k\in I_k}\left(\sum_{i'_k \in I_k} T^*_{\pi,i'_k,i_k}-\hat U_{\lambda,\pi,i'_k}\hat U_{\lambda,\pi,i_k}^t\right)^2 \\
             \le     \frac{(1+\epsilon) \ \sqrt{1+\mu (K-1)}}{1-\mu(K-1)} 
             \min_{\tilde U \in \mathbb O_{n,R,+}, \ M\tilde U \tilde U^t=M}\quad \Vert M(T^*-\tilde U\tilde U^{t})\Vert_F \\ +\sqrt{\frac{1}
      {\exp \left( \frac{\left(\sqrt{c_U^2-c_O^2}-\sqrt{dR} \right)^2}{2} \right) - 1}}  +\sqrt{(dR-\frac12 (R^2+R)) \log(\rho^{-1})+\log(c^{-1})} \\
      +(2+\epsilon)\nu_{\max}.
    \end{multline*}
    with\footnote{Notice that $t_{\min}$ needs to be sufficiently smaller that $2/e$ in order for the term $K$ to become small for $n$ sufficiently large.} 
    \begin{align*}
    t_{\min}  & = \frac{\left(\frac{\nu_{\min}}{\sigma_{\min}} + 4 \epsilon \sqrt{nd+u}\right)^2}{n (n-R)}
    \end{align*}
    and
    \begin{align*}t_{\max}= \left(\frac{\nu_{\max}}{\sigma_{\max}} - 4 \epsilon \sqrt{nd+u}\right)^2-\sqrt{n (n-R)},
    \end{align*}
  \end{theorem}
This theorem shows that a bound on the difference of the cluster matrices can be obtained when the matrix of means is sufficiently incoherent. Notice that this bound is not exactly component-wise, but considers a sum over clusters, which is perfectly relevant because the matrix $M$ does not distinguish between points in the same cluster. As expected, the smaller the coherence $\mu$, the better the oracle bound.

    The proof of Theorem \ref{incoh} is deferred to the supplementary material.
  

  \section{A Langevin sampler}\label{sec:algo}
  In this section, we present a Langevin sampler for our estimator $\hat U_\lambda,\hat O_\lambda$ \footnote{Notation-wise, we will identify the Stiefel manifold with the set of matrices whose first $R$ columns form an orthonormal family and the remaining $n-R$ columns are set to zero.}.
  Langevin-type samplers were first proposed by 
  \cite{grenander1983tutorial}, \cite{grenander1994representations}, \cite{roberts1996exponential}, and have attracted a lot a attention lately in the statistical learning community \cite{dalalyan2017theoretical,durmus2017nonasymptotic,brosse2018promises}.
  
  \subsection{Computing the gradient on the Stiefel manifold}
  We start with some preliminary material about gradient computation on the Stiefel manifold from \cite{edelman1998geometry}. The Stiefel manifold can be interpreted as the set of equivalence classes
  \begin{align*}
  [O] & = 
  \left\{
  \begin{bmatrix}
  O 
  \begin{bmatrix}
  I_R & 0 \\
  0 & O'
  \end{bmatrix},
  \quad \mathrm{with} \quad  O' \in \mathbb O_{d-R}
  \end{bmatrix}
  \right\}.
  \end{align*}
  As can easily be deduced from this quotient representation of the Stiefel manifold, the tangent space to the Stiefel manifold at a point $O$ is 
  \begin{multline*}
  T_O(\mathbb O_{d,R}) =  
  \Bigg\{O 
  \begin{bmatrix}
  A & -B^t \\
  B & 0 
  \end{bmatrix},
  \quad \mathrm{with} \quad A \in \mathbb R^{R\times R} \ \mathrm{skew \ symmetric}\Bigg\}.
  \end{multline*}
  The canonical metric at a point $O$ is given by  
  \begin{align*}
  g_c & = \mathrm{trace}\left(\Delta^t\left(I-\frac12 OO^t\right)\Delta\right).
  \end{align*}
  For $\Delta \in T_O(\mathbb O_{d,R})$, the exponential map is given by 
  $
  O(t)  = O e^{t\Delta} I_{d,R}.
  $
  The gradient at $O$ of a function $f$ defined on the Stiefel manifold $\mathbb O_{d,R}$ is given by \footnote{This formula can be obtained using differentiation along the geodesic defined by the exponential map in the direction $\Delta$, for all $\Delta \in T_O(\mathbb O_{d,R})$.} 
  \begin{align}
  \nabla f(O) & = f_O-Of_O^tO,
  \end{align}
  where 
  $
  f_O(i,i')  = \frac{\partial f}{\partial\ O_{i,i'}} 
  $ for any $i$, $i'=1,\ldots,n$.
  
  \subsection{The alternating Langevin sampler}
  
  The alternating Langevin sampler is described as follows. It consists in alternating between perturbed gradient steps in the matrix variable $O$ on the Stiefel manifold 
  and perturbed gradient steps in the matrix variable $U$.
  
For clarity of the exposition, we give the formula for the gradient of $\Vert X-XUU^t\Vert_F^2$ as a function of $U$:
  \begin{align*}
  \nabla \left(\Vert X-X \cdot \cdot^t\Vert_F^2\right)_U  = \left((X-XUU^t)^t X+ X^t(X-XUU^t)\right) U.
  \end{align*}
Following \cite{brosse2018promises}, we propose the following algorithm.  
  
  \begin{algorithm}[h]
    \SetAlgoLined
    \KwResult{A sample $\hat U_\lambda$ of the quasi-posterior distribution}
    initialise $U^{(0)}=O^{(0)}$\\
    \For{$\ell=1$}
    {
      \begin{align*}
      O^{(\ell+1)} & = \exp \Bigg(O^{(\ell)},-h \ \Big(\mathrm{sign}(O^{(\ell)}) \odot\left(U^{(\ell)}-\vert O^{(\ell)}\vert\right) \\
      & \hspace{-.2cm} -O^{(\ell)} 
      \left(\mathrm{sign}(O^{(\ell)}) \odot \left(U^{(\ell)}-\vert O^{(\ell)}\vert\right)\right)^tO^{(\ell)}\\
      & +\sqrt{2\ h} \ Z_O^{(\ell)}\Big) \Bigg) \\    
      U^{(\ell+1)} & =  U^{(\ell)}-h \ \Bigg(-\bigg((X-XU^{(\ell)}U^{(\ell)^t})^t X \\
      & + X^t(X-XU^{(\ell)}U^{(\ell)^t})\bigg)U^{(\ell)} \\
      & \hspace{-.2cm} +\frac1{\mu^2}\ \left( U^{(\ell)}-\vert O^{(\ell+1)}\vert\right)\Bigg)+\sqrt{2 \ h} \ Z_U^{(\ell)}.
      \end{align*} 
    }
    \caption{The Langevin sampler}
  \end{algorithm}
  
  In this algorithm the exponential function $\exp(O,H)$ at $O$ is given by \cite[][Eq. 2.45]{edelman1998geometry} using different notation.

  \section{Numerical experiment}\label{sec:simus}
\begin{figure}
      \centering
      \includegraphics[width=10cm]{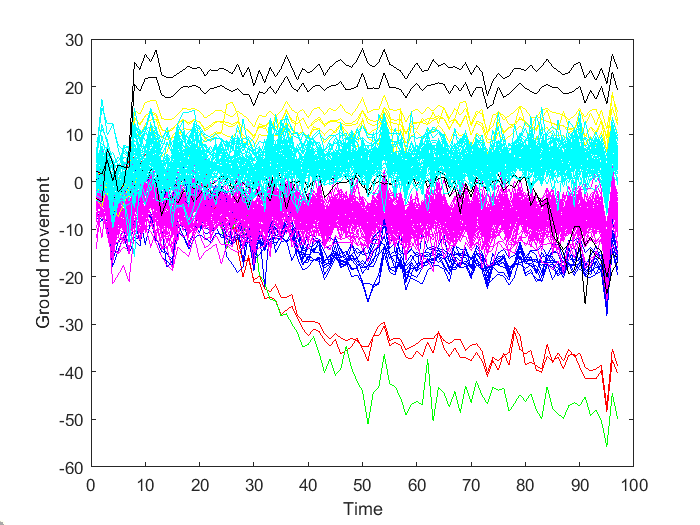}
      \caption{Clustered satellite data representing ground movements}
      \label{fig:sat}
  \end{figure}

  We provide some illustrations of the performance of our approach to clustering using a simple two component Gaussian Mixture and real satellite data. 
We  performed an experiment with real high dimensional (satellite time series) data and one sample from the posterior after 300 iterations gave the result shown in Figure \ref{fig:sat}.
  
\section{Proof of Theorem 2.1}\label{sec:proof}
  
  We break down the proof in the following successive elementary steps.
  
  \subsection{Initial PAC-Bayesian bound}
  
  \begin{theorem}[adapted from \citealp{dalalyan2008aggregation}]
    For $\lambda \le 1/4$, 
    \begin{align}
    \mathbb E \left[ \Vert X-X\hat U_{\lambda} \hat U_{\lambda}^t\Vert_F^2\right] & \le \inf_{\rho} \ 
    \left\{\int \ \Vert X-X\hat U_{\lambda} \hat U_{\lambda}^t\Vert_F^2 \ \rho(U) \ dU \right\}+\frac{KL(\rho,\pi)}{\lambda}
    \label{daltsy}
    \end{align}
    where the infimum is taken over all probability measures $\rho$ which are absolutely continuous with respect to $\pi$. Here $KL$ denotes the Kullback-Leibler divergence.
  \end{theorem}
  
  \subsection{Bounding the integral part}
  In order to bound the integral part in the bound given in \eqref{daltsy}, we define for any $R$ and any matrix $U^0\in \mathcal U_R \cap \mathbb O_{n,R,+}$ for any $c_U, \ c_O\in (0,1]$, the measure 
  \begin{align*}
  \rho_{R,U^0,c_U,c_O}(U,O) & = \frac{1_{\Vert U-U^0\Vert_F \le c_U, \ \Vert \vert O\vert-U^{0}\Vert_F \le c_O} \quad \pi_{U,O}(U,O)}{
    \pi_{U,O}\left(\Vert U-U^0\Vert_F \le c_U, \ \Vert \vert O\vert-U^{0}\Vert_F \le c_O\right)}.
  \end{align*}
  Define $c=(c_U,c_O)$. Using these distributions we will be able to prove the following bound. 
  \begin{lemma}
    We have 
    \begin{align*}
    \int \Vert X-XUU^t \Vert_F^2 \rho_{R,U^0,c_U,c_O}(U,O) \ dU dO   \le 
    \left(\Vert X-XU^0U^{0^t}\Vert_F + c_U  \left(2+c_U\right) \ \left(\Vert M\Vert+\Vert E \Vert\right)\right)^2.
    \end{align*}
    \label{firsterm}
  \end{lemma}
  \begin{proof}
    Note that 
    \begin{align*}
    \int \Vert X-XUU^t \Vert_F^2 \rho_{R,U^0,c_U,c_O}(U,O) \ dU dO  & \\
    & \hspace{-4cm} = \int \Bigg(\Vert X-XU^0U^{0^t} \Vert_F^2 + 2 \langle X(UU^t-U^0U^{0^t}, X-XU^0U^{0^t} \rangle \\
    & \hspace{-4cm} + \Vert XUU^t-XU^0U^{0^t} \Vert_F^2\Bigg) \ \rho_{R,U^0,c_U,c_O}(U,O) \ dU dO 
    \end{align*}
    and thus, 
    \begin{align*}
    \int \Vert X-XUU^t \Vert_F^2 \quad \rho_{R,U^0,c_U,c_O}(U,O) \ dU dO & \\
    & \hspace{-6cm} = \int  \Big( \Vert X-XU^{0}U^{0^t}\Vert_F^2+ 2 \langle X(UU^t-U^0U^{0^t}), X-XU^{0}U^{0^t}\rangle \\ 
    & \hspace{-4cm}
    + \Vert XUU^t-XU^0U^{0^t} \Vert_F^2 \Big)\ \rho_{R,U^0,c_U,c_O}(U,O) \ dU dO.
    \end{align*}
    By the Cauchy-Schwartz inequality, we get 
    \begin{align*}
    \int \Vert X-XUU^t \Vert_F^2 \quad \rho_{R,U^0,c_U,c_O}(U,O) \ dU dO & = \Vert X-XU^{0}U^{0^t}\Vert_F^2  \\
    & \hspace{-6cm}  + \Vert X-XU^{0}U^{0^t}\Vert_F \ \int   2 \ \Vert X(UU^t-U^0U^{0^t})\Vert_F \ \rho_{R,U^0,c_U,c_O}(U,O) \ dU dO  \\
    & \hspace{-6cm}  +  \int \Vert X(UU^t-U^0U^{0^t}) \Vert_F^2 \ \rho_{R,U^0,c_U,c_O}(U,O) \ dU dO  
    \end{align*}     
    Note further that since $(U,O)$ must belong to the support of $\rho_{R,U^0,c_U,c_O}$, we have 
    \begin{align*}
    \Vert X(UU^t-U^0U^{0^t}) \Vert_F & = \left\Vert X\left(U(U-U^0)^t+(U-U^0)U^{0^t}\right)\right\Vert_F \\
    & \le \left\Vert XU(U-U^0)^t\right\Vert_F+\left\Vert X(U-U^0)U^{0^t}\right\Vert_F.
    \end{align*}
    Now, on the one hand, we have 
    \begin{align*}
    \left\Vert XU(U-U^0)^t\right\Vert_F^2 & \le \Vert XU \Vert^2 \ \Vert U-U^0\Vert_F^2 \nonumber \\
    & \le \left(\Vert MU^{0}\Vert+\Vert EU^{0}\Vert +(\Vert M\Vert +\Vert E\Vert)\Vert U-U^{0}\Vert\right)^2 \ \Vert U-U^0\Vert_F^2 
    \end{align*}
    and since the columns of $U^0$ are orthonormal, we have $\Vert U^0\Vert=1$, which gives 
    \begin{align*}
    \left\Vert XU(U-U^0)^t\right\Vert_F & \le \left(\Vert M\Vert+\Vert E \Vert\right)  \left(1+\Vert U-U^{0}\Vert_F\right) \ \Vert U-U^0\Vert_F.
    \end{align*}
    On the other hand, 
    \begin{align*}
    \left\Vert X(U-U^0)U^{0^t}\right\Vert_F^2 & \le \Vert X \Vert^2  \left\Vert (U-U^0)U^{0^t} \right\Vert_F^2 
     \le \Vert X \Vert^2 \ \Vert U^{0} \Vert^2 \left\Vert (U-U^0) \right\Vert_F^2 
    \end{align*}
    and using again that $\Vert U^{0}\Vert =1$, 
    \begin{align*}
    \left\Vert X(U-U^0)U^{0^t}\right\Vert_F & \le \Vert X \Vert \ \left\Vert (U-U^0) \right\Vert_F.
    \end{align*}
    From this, we easily deduce that 
    \begin{align*}
    \int \Vert X-X(UU^t)\Vert_F \ \rho_{R,U^0,c_U,c_O}(U,O) \ dU dO   &\le \left(\Vert M\Vert+\Vert E \Vert\right)  \left((1+c_U) \ c_U +\ c_U\right),  \\
    &  \le c_U  \left(2+c_U\right) \ \left(\Vert M\Vert+\Vert E \Vert\right) ,         
    \end{align*}
    and 
    \begin{align*}
    \int \Vert X-X(UU^t)\Vert_F^2 \ \rho_{R,U^0,c_U,c_O}(U,O) \ dU dO  &   \le c_U^2 \left(2+c_U\right)^2 \ \left(\Vert M\Vert+\Vert E \Vert\right)^2  
    \end{align*}
    which completes the proof. 
  \end{proof}
  \subsection{Upper bound on the Kullback-Leibler divergence}

  \begin{lemma}
    We have 
    \begin{align*}
    KL \left(\rho_{R,U^0,c},\pi \right) & \le \frac{1}
    {\exp \left( \frac{\left(\frac1{\mu} \ \sqrt{c_U^2-c_O^2}-\sqrt{nR} \right)^2}{2} \right)-1}   + \left(nR-\frac12 (R^2+R)\right) \log(\rho^{-1})+\log(c^{-1}),
    \end{align*}
    for some $\rho,c>0$ sufficiently small.
    \label{scdterm}
  \end{lemma}
  \begin{proof} 
    By definition, 
    \begin{align*}
    KL(\rho_{R,U^0,c_U,c_O},\pi) & = \int \rho_{R,U^0,c_U,c_O}(U,O) \log\left(\frac{\rho_{R,U^0,c_U,c_O}(U,O)}{\pi_{U,O}(U,O)}\right) \ dU dO, \\
    & =  \log\left(\frac{1}{\int 1_{\Vert U-U^0 \Vert_F\le c_U, \ \Vert \vert O\vert-U^{0} \Vert_F\le c_O} \ \pi_{U,O}(U,O) \ dU dO }\right) \ dU dO.
    \end{align*}
    We then have 
    \begin{align*}
    \pi_{U,O}\left(\Vert U-U^0\Vert_F^2 \le c_U^2, \quad 
    \Vert \vert O\vert -U^0 \Vert_F^2 \le c_O^2\right) & \\
    \\
    & \hspace{-7cm} \ge   \pi_{U,O}\left(\Vert U-\vert O\vert \Vert_F^2 \le c_U^2-c_0^2, \quad 
    \Vert \vert O\vert -U^0 \Vert_F^2 \le c_O^2\right) & \\
    & \hspace{-7cm} = 
    \int_{\Vert \vert O\vert -U^0 \Vert_F^2 \le c_O^2} \left( \int 1_{\Vert U-\vert O\vert  \Vert_F^2 \le c_U^2-c_O^2} \quad  \pi_{U\mid O}(U) dU \right)\pi_O(O) dO \\
    & \hspace{-7cm} = \int_{\Vert \vert O\vert -U^0 \Vert_F^2 \le c_O^2} \ \pi_{U\mid O} \left( \sum_{i=1}^d\sum_{r=1}^R (U_{i,r}-\vert O_{i,r}\vert )^2 \le c_U^2-c_O^2\right) \ \pi_O(O) dO. 
    \end{align*}
    As long as $c_U^2\ge  dR$, the inner probability can be bounded as follows \citep[see equation 7.50 in][]{massart2007concentration}:
    \begin{align*}
    \pi_{U\mid O} \left( \sqrt{ \sum_{i=1}^d\sum_{r=1}^R (U_{i,r}-\vert O_{i,r}\vert )^2} \le  \sqrt{c_U^2-c_O^2}\right) & \ge 1-\exp \left( -\frac{\left( \frac{1}{\mu} \ \sqrt{c_U^2-c_O^2}-\sqrt{nR} \right)^2}{2} \right).
    \end{align*}
    From this last inequality, we get 
    \begin{align}
    \pi_{U,O}\left(\Vert U_{i,r}-U^{0}_{i,r}\Vert_F^2 \le c_U^2, \quad 
    \Vert \vert O\vert -U^0 \Vert_F^2 \le c_O^2\right) & \nonumber \\
    & \hspace{-7cm}\ge \left(1-\exp \left( -\frac{\left(\frac{1}{\mu} \ \sqrt{c_U^2-c_O^2}-\sqrt{nR} \right)^2}{2} \right)\right) \ \pi_O\left(\Vert \vert O\vert -U^0 \Vert_F^2 \le c_O^2\right). 
    \label{titi}
    \end{align}
    We now use the elementary inequality $\log(1+x) \ge x/(1+x)$ for $x\in (-1,+\infty)$, and a lower bound on $\pi_O\left(\Vert \vert O\vert -U^0 \Vert_F^2 \le c_O^2\right)$ \citep[given by][]{leeruymgaart96}. Therefore
    \begin{align*}
    \pi_O\left(\Vert \vert O\vert -U^0 \Vert_F^2 \le c_O^2\right) & \ge c \ \rho^{nR-\frac12 (R^2+R)}
    \end{align*} 
    for $c$ and $\rho$ sufficiently small. 
    From these two inequalities, we end the proof:
    \begin{align*}
    \log\left(\pi_{U\mid O} \left( \sqrt{ \sum_{i=1}^d\sum_{r=1}^R (U_{i,r}-\vert O_{i,r}\vert )^2} \le  \sqrt{c_U^2-c_O^2}\right)\right) & \ge 
    \frac{-1}{\exp \left( \frac{\left(\frac{1}{\mu} \ \sqrt{c_U^2-c_O^2} -\sqrt{dR} \right)^2}{2} \right)-1}\nonumber \\
    & \hspace{-4cm}
    + \left(nR-\frac12 (R^2+R)\right) \log(\rho)+\log(c).
    \end{align*}    
  \end{proof}
  
  \subsection{Combining the lemm\ae}
  Combining the results of the previous lemm\ae, we get the following proposition.
  \begin{proposition}
    \label{propop}
    Let 
    \begin{align}
        \label{epscu}
        \nu & = \min_{\tilde U \in \mathbb O_{n,R,+}, \ M\tilde U \tilde U^t=M}\quad \Vert E(I-\tilde U \tilde U^t)\Vert_F.
    \end{align} 
    Then, for any $c_U$ such that 
    $
    c_U  \left(2+c_U\right)  \le \epsilon \frac{\nu}{\left(\Vert M\Vert+\Vert E \Vert\right)},
    $
    we have 
    \begin{align*}
    \mathbb E \left[\left\Vert X- X\hat U_{\lambda}\hat U_{\lambda}^t\right\Vert_F^2\right] & \le 
    (1+\epsilon)^2  \Vert X-XU^0U^{0^t}\Vert_F^2 \\
    &\hspace{-2cm} +\frac{1}
    {\exp \left( \frac{\left(\frac{1}{\mu} \ \sqrt{c_U^2-c_O^2}-\sqrt{nR} \right)^2}{2} \right)-1}  + (nR-\frac12 (R^2+R)) \log(\rho^{-1})+\log(c^{-1}). 
    \end{align*} 
  \end{proposition}
  \begin{proof}
    Let $\tilde U$ be a minimiser in the right hand-side in \eqref{epscu}. Since 
    \begin{align*}
    \Vert X-X\tilde U\tilde U^{t}\Vert_F & = \Vert M+E-M\tilde U\tilde U^{t}-E\tilde U\tilde U^{t}\Vert_F 
    \end{align*}
    and since $M=M\tilde U\tilde U^{t}$, we get 
    $
    \Vert X-X\tilde U\tilde U^{t}\Vert_F  = \Vert E(I-\tilde U\tilde U^{t})\Vert_F.
    $
    As 
    \begin{align*}
    \Vert E(I-\tilde U\tilde U^{t})\Vert_F & \le \Vert E(I- U^0 U^{0^t})\Vert_F
    \end{align*}
    for all $U^0\in \mathbb O_{n,r,+} \cup \{U \mid M U U^t=M\}$, the claim then follows from combining the results of Lemma \ref{firsterm} and Lemma \ref{scdterm} above, and taking $c_U$ such that (3) in the main text holds.
  \end{proof}

  \subsection{Assembling the elements}
  We have that 
  \begin{align*}
  \left\Vert X- X\hat U_{\lambda}\hat U_{\lambda}^t\right\Vert_F^2 & \ge  \left(\left\Vert M- M\hat U_{\lambda}\hat U_{\lambda}^t\right\Vert_F- \nu_{\max}
  \right)^2  =  \left(\left\Vert M \left(U^0U^{0^t}-\hat U_{\lambda}\hat U_{\lambda}^t\right)\right\Vert_F- \nu_{\max}
  \right)^2
  \end{align*}
  Using Jensen's inequality gives 
  \begin{align*}
  \mathbb E \left[\left\Vert X- X\hat U_{\lambda}\hat U_{\lambda}^t\right\Vert_F^2\right] & \ge 
  \mathbb E \left[\left\Vert X- X\hat U_{\lambda}\hat U_{\lambda}^t\right\Vert_F\right]^2  \ge \mathbb E \left[\left\Vert M \left(T^*-\hat U_{\lambda}\hat U_{\lambda}^t\right)\right\Vert_F- \nu_{\max}\right]^2
  \end{align*}
  which, combined with Proposition \ref{propop} gives 
  \begin{align*}
  \mathbb E \left[\left\Vert M \left(T^*-\hat U_{\lambda}\hat U_{\lambda}^t\right)\right\Vert_F\right] & \le  (1+\epsilon)  \Vert X-XU^0U^{0^t}\Vert_F \nonumber \\
  \nonumber \\
  & \hspace{-4cm}+\sqrt{\frac{1}
    {\exp \left( \frac{\left(\frac{1}{\mu} \ \sqrt{c_U^2-c_O^2}-\sqrt{nR} \right)^2}{2} \right)-1}} +\sqrt{(nR-\frac12 (R^2+R)) \log(\rho^{-1})+\log(c^{-1})}
    +\nu_{\max}.
  \end{align*} 
  Given that 
  \begin{align*}
  \Vert X-XU^0U^{0^t} \Vert_F & \le \Vert M - M U^0U^{0^t} \Vert_F + \nu_{\max}
  \end{align*}
  this completes the proof of Theorem 2.1, since $U^0$ is any matrix satisfying the constraints.

  \section{Proof of Theorem 2.2 }\label{sec:appendix}
  \subsection{Control of $\mathbb P \left(\min_{U \in \mathbb O_{n,R}} \ \Vert E(I-UU^t)\Vert_F \le \nu_{\min} \right)$}
  The covering number of the Stiefel manifold in the operator norm was computed in \cite{hinrichs2017entropy} and is given by 
  \begin{align}
  N(\mathbb O_{n,R},\Vert \cdot \Vert,\epsilon) & \le \left(\frac{c}{\epsilon}\right)^{nR-R(R+1)/2}.
  \end{align}
  Let $\mathcal N_\epsilon$ denote an $\epsilon$-net in the operator norm for the Stiefel manifold with cardinality $N(\mathbb O_{n,R},\Vert \cdot\Vert,\epsilon)$. 
  For any $U \in \mathbb O_{n,R}$, let $U^\sharp$ denote the closest matrix in $\mathcal N_\epsilon$ to $U$. Then, we have 
  \begin{align}
  \Vert E(I-UU^{t})\Vert_F & = \Vert E(I-(U^\sharp+(U-U^\sharp)) (U^\sharp+(U-U^\sharp))^{t}\Vert_F \nonumber \\
  & \ge \Vert E(I-U^\sharp U^{\sharp^t})\Vert_F -\Vert E U^\sharp(U-U^\sharp)^t)\Vert_F -\Vert E (U-U^{\sharp})U^{\sharp^t})\Vert_F \nonumber \\
  & \ge \Vert E(I-U^\sharp U^{\sharp^t})\Vert_F -\Vert (U-U^\sharp)\Vert \Vert E U^\sharp\Vert_F -\Vert E (U-U^{\sharp})U^{\sharp^t}\Vert_F \label{tititi}
  \end{align}
  Moreover, 
  \begin{align*}
  \Vert E (U-U^{\sharp})U^{\sharp^t}\Vert_F^2 & = 
  \mathrm{trace}(E(U-U^{\sharp})U^{\sharp^t}U^{\sharp}(U-U^{\sharp})^tE^t) \\
  & = \mathrm{trace}(E(U-U^{\sharp})(U-U^{\sharp})^tE^t) \\
  & = \Vert E(U-U^\sharp) \Vert_F^2 \\
  & \le \Vert U-U^\sharp\Vert^2 \  \Vert E\Vert_F^2
  \end{align*}
  Combining this last inequality with \eqref{tititi}, we get 
  \begin{align}
  \Vert E(I-UU^{t})\Vert_F  
  & \ge \Vert E(I-U^\sharp U^{\sharp^t})\Vert_F -\epsilon \left( \Vert E U^\sharp\Vert_F + 
  \Vert E\Vert_F\right) \nonumber \\
  & \le \Vert E(I-U^\sharp U^{\sharp^t})\Vert_F -2 \ \epsilon \Vert E\Vert_F \nonumber  
  \end{align}
  Therefore, we obtain that  
  \begin{align*}
  \mathbb P \left(\min_{U \in \mathbb O_{n,R}} \ \Vert E(I-UU^t)\Vert_F \le \nu_{\min} \right) & \le 
  \mathbb P \left(\min_{U \in \mathcal N_\epsilon} \ \Vert E(I-UU^t)\Vert_F \le \nu_{\min}+2\epsilon \Vert E\Vert_F \right) \\
  & \le 
  \mathbb P \left(\min_{U \in \mathcal N_\epsilon} \ \Vert E(I-U^\sharp U^{\sharp^t})\Vert_F \le \nu_{\min}+4\epsilon \Vert E\Vert_F \right)
  \end{align*}
  where, for any $U$ in $\mathcal N_\epsilon$, $U^\sharp$ will denote the projection in operator norm of $U$ onto $\mathbb O_{n,R}$. Moreover
  \begin{align*}
  \mathbb P \left(\min_{U \in \mathbb O_{n,R}} \ \Vert E(I-UU^t)\Vert_F \le \nu_{\min} \right) & \le \mathbb P \left(\min_{U \in \mathcal N_\epsilon} \ \Vert E(I-U^\sharp U^{\sharp^t})\Vert_F \le \nu_{\min}+4\epsilon \Vert E\Vert_F, \ \Vert E\Vert_F\le \eta \right) \\
  & \hspace{.3cm}+\mathbb P \left(\min_{U \in \mathcal N_\epsilon} \ \Vert E(I-U^\sharp U^{\sharp^t})\Vert_F \le v+4\epsilon \Vert E\Vert_F, \ \Vert E\Vert_F > \eta\right)\\
  & \le \mathbb P \left(\min_{U \in \mathcal N_\epsilon} \ \Vert E(I-U^\sharp U^{\sharp^t})\Vert_F \le \nu_{\min}+4\epsilon \eta \right) \\
  & \hspace{.3cm}+\mathbb P \left( \Vert E\Vert_F > \eta\right)
  \end{align*}
  Since $E$ is i.i.d. Gaussian with minimum one-dimensional variance $\sigma_{\min}^2$, for any $U^\sharp \in \mathbb O_{n,R}$,
  the lower tail of $\sigma_{\min}^{-1}\Vert E(I-U^\sharp U^{\sharp^t})\Vert_F^2$ is dominated by the lower tail of a $\chi^2(n(n-R))$ distribution and therefore, as recalled in \citet[][Lemma B1]{chretien2014sparse} 
  \begin{align}
  \mathbb P\left(\Vert E(I-U^\sharp U^{\sharp^t})\Vert_F \le \sigma \ \sqrt{t \ n(n-R)}\right) & \le \frac{2}{\sqrt{\pi \ n(n-R)}} \ \left(t \ e/2 \right)^{n(n-R)/4}.  
  \end{align}
  Let us now tune $t$ and $\eta$. On the one hand, by \cite{boucheron2013concentration}, we have that 
  \begin{align*}
  \mathbb P \left(\Vert E\Vert_F \ge  \sigma \sqrt{d n+u}\right) & \le \exp (-nu^2/8).
  \end{align*}
  Thus, we will choose $\eta = \sigma \sqrt{d n+u}$.
  On the other hand, we will take $t$ such that 
  \begin{align*}
  \sigma \sqrt{t n (d-R)} & = \nu_{\min} + 4 \epsilon \eta 
  \end{align*}
  i.e. 
  \begin{align*}
  t  & = \frac{\left(\frac{\nu_{\min}}{\sigma} + 4 \epsilon \sqrt{nd+u}\right)^2}{n (n-R)} 
  \end{align*}
  Therefore, using the union bound we get 
  \begin{align*}
  \mathbb P \left(\min_{U \in \mathcal N_\epsilon} \ \Vert E(I-U^\sharp U^{\sharp^t})\Vert_F \le \nu_{\min} \right) & \le  \frac{2 \ N(\mathbb O_{n,R},\Vert \cdot\Vert,\epsilon)}
  {\sqrt{\pi \ n(n-R)}} \ \left(t \ e/2 \right)^{n(n-R)/4}.
  \end{align*}
  
  
  \subsection{Control of $\mathbb P \left(\max_{U \in \mathbb O_{n,R}} \ \Vert E(I-UU^t)\Vert_F \ge \nu_{\max} \right)$}
  The same strategy applies, with slight modifications. We consider the same $\epsilon$-net as in the previous subsection. 
  We first easily get 
  \begin{align}
  \Vert E(I-UU^{t})\Vert_F & \le  \Vert E(I-U^\sharp U^{\sharp^t})\Vert_F^2+ 2 \ \epsilon \Vert E\Vert_F \nonumber  
  \end{align}
  Therefore,   
  \begin{align*}
  \mathbb P \left(\max_{U \in \mathbb O_{n,R}} \ \Vert E(I-UU^t)\Vert_F \ge \nu_{\max} \right) & \le 
  \mathbb P \left(\max_{U \in \mathcal N_\epsilon} \ \Vert E(I-UU^t)\Vert_F \ge \nu_{\max}-2\epsilon \Vert E\Vert_F \right) \\
  & \le 
  \mathbb P \left(\max_{U \in \mathcal N_\epsilon} \ \Vert E(I-U^\sharp U^{\sharp^t})\Vert_F \le \nu_{\max}-4\epsilon \Vert E\Vert_F \right)
  \end{align*}
  where, for any $U$ in $\mathcal N_\epsilon$, $U^\sharp$ again denotes the projection in operator norm of $U$ onto $\mathbb O_{n,R}$. We then get 
  \begin{align*}
  \mathbb P \left(\max_{U \in \mathbb O_{n,R}} \ \Vert E(I-UU^t)\Vert_F \le \nu_{\max} \right) 
  & \le \mathbb P \left(\max_{U \in \mathcal N_\epsilon} \ \Vert E(I-U^\sharp U^{\sharp^t})\Vert_F \le \nu_{\max}-4\epsilon \eta \right) \\
  & \hspace{.3cm}+\mathbb P \left( \Vert E\Vert_F > \eta\right)
  \end{align*}
  Since $E$ is i.i.d. Gaussian with maximum one-dimensional variance $\sigma_{\max}^2$, the upper tail of $\sigma_{\max}^{-1}\Vert E(I-U^\sharp U^{\sharp^t}) \Vert_F^2$ is dominated by that of a $\chi^2(n(n-R))$ distribution for any $U^\sharp$ in $\mathbb O_{n,R}$, and therefore, as recalled in \citet[][Lemma B1]{chretien2014sparse} 
  \begin{align}
  \mathbb P\left(\Vert E(I-U^\sharp U^{\sharp^t})\Vert_F \ge \sigma_{\max} \ \left(\sqrt{n(n-R)}+\sqrt{2t}\right)\right) & \le \exp(-t). 
  \end{align}
  We can now tune $t$ and $\eta$. Recall that for all $u>0$, we have 
  \begin{align*}
  \mathbb P \left(\Vert E\Vert_F \ge  \sigma \sqrt{d n+u}\right) & \le \exp (-nu^2/8).
  \end{align*}
  Thus, we will choose as before $\eta = \sigma_{\max} \sqrt{d n+u}$.
  On the other hand, we will take $t$ such that 
  \begin{align*}
  \sigma_{\max}\left( \sqrt{2t}+ \sqrt{n (n-R)} \right) & = \nu_{\max} - 4 \epsilon \eta 
  \end{align*}
  i.e. 
  \begin{align*}
  t  & = \left(\frac{v}{\sigma_{\max}} - 4 \epsilon \sqrt{nd+u}\right)^2-\sqrt{n (d-R)}. 
  \end{align*}
  Therefore, using the union bound we get 
  \begin{align*}
  \mathbb P \left(\min_{U \in \mathcal N_\epsilon} \ \Vert E(I-U^\sharp U^{\sharp^t})\Vert_F \le v \right) & \le  N(\mathbb O_{n,R},\Vert \cdot\Vert,\epsilon) \ \exp(-t).
  \end{align*}

  \subsection{Control of $\Vert E\Vert$}
  We will also need to control $\Vert E\Vert$. Using \citet[][Section 4.4]{vershynin2018high}, we obtain 
  \begin{align*}
  \Vert E\Vert \le \sigma \left(\sqrt{n}+2\sqrt{d}\right) 
  \end{align*} 
  with probability at least $1-\exp(-d)$.
  \subsection{Control of $\Vert M\Vert$}
  Finally, we need to compute $\Vert M\Vert$ as a function of the coherence $\mu$. Using the Gershgorin bound, we easily obtain  
  \begin{align}
      \Vert M\Vert =\sqrt{\Vert M^tM-I\Vert+1} & \le \sqrt{\left(\max_{k=1}^K \ n_k\right) \ \mu(K-1)+1}
  \end{align} 
  
  \subsection{End of the proof}
  Let us now study $\Vert M(T^*\hat U_\lambda \hat U_\lambda^t)\Vert_F^2$. Let 
  \begin{align*}
      \Upsilon & = [\mu_1,\ldots,\mu_K].
  \end{align*}
  Let $\Upsilon^\dagger$ denote the pseudo inverse of $\Upsilon$, i.e. 
  \begin{align*}
      \Upsilon^\dagger & = (\Upsilon^t\Upsilon)^{-1}\Upsilon^t.
  \end{align*}
  In particular, let $\pi$ denote a permutation which orders the data cluster wise, i.e. all data from cluster 1, followed by all data from cluster 2, \ldots, all data from cluster $K$ and let $M_\pi$ denote the matrix whose columns are permuted by $\pi$. Then, 
  \begin{align*}
      \Upsilon^\dagger M_\pi & = 
      \begin{bmatrix}
      1_{n_1}^t & 0 & \cdots & \cdots & 0 \\
      0 & 1_{n_2}^t & \cdots & \cdots & 0 \\
      \vdots &  0    &        &        & \vdots \\
      \vdots &  \vdots    &        &        & \vdots \\
      0 & 0 & \cdots & 0 & 1_{n_K}^t
      \end{bmatrix}
  \end{align*}
  Denote by $T^*_\pi$ the matrix obtained from $T^*$ after reordering its rows and columns using $\pi$, and $\hat U_{\lambda,\pi}$ denote the matrix obtained from $\hat U_{\lambda}$ after reordering its row using $\pi$.    
  \begin{align*}
      \left\Vert \Upsilon^\dagger M_\pi \left(T^*_\pi-\hat U_{\lambda,\pi} U_{\lambda,\pi}\right)\right\Vert_F^2 & = 
      \sum_{k=1}^K \sum_{i_k\in I_k}\left(\sum_{i'_k \in I_k} T^*_{\pi,i'_k,i_k}-\hat U_{\lambda,\pi,i'_k}\hat U_{\lambda,\pi,i_k}^t\right)^2. 
  \end{align*}
  On the other hand,
  \begin{align*}
      \sigma_{\max} \left(\Upsilon^\dagger\right) & \le \sigma_{\min} \left( \Upsilon^t\Upsilon\right)^{-1} \sigma_{\max}\left( \Upsilon^t \right) \ge \frac{\sqrt{1+\mu (K-1)}}{1-\mu(K-1)}
  \end{align*}
  and we get that 
  \begin{align*}
      \left\Vert \Upsilon^\dagger M_\pi \left(T^*_\pi-\hat U_{\lambda,\pi} U_{\lambda,\pi}\right)\right\Vert_F^2 
      & \le \frac{\sqrt{1+\mu (K-1)}}{1-\mu(K-1)} \ \left\Vert M_\pi \left(T^*_\pi-\hat U_{\lambda,\pi} U_{\lambda,\pi}\right)\right\Vert_F^2
  \end{align*}
  and the result follows.

\bibliographystyle{plainnat}
  \bibliography{bib}

\end{document}